\documentclass{article}
\usepackage{spconf}
\usepackage[utf8]{inputenc} % allow utf-8 input
\usepackage[T1]{fontenc}    % use 8-bit T1 fonts
\usepackage{hyperref}       % hyperlinks
\usepackage{url}            % simple URL typesetting
\usepackage{booktabs}       % professional-quality tables
\usepackage{amsfonts}       % blackboard math symbols
\usepackage{nicefrac}       % compact symbols for 1/2, etc.
\usepackage{microtype}      % microtypography
\usepackage[algoruled,boxed,lined]{algorithm2e}
\usepackage{graphicx}
\usepackage{pifont}% http://ctan.org/pkg/pifont
\usepackage[tight,footnotesize]{subfigure}
\usepackage{amsmath} % assumes amsmath package installed
\usepackage{amssymb}  % assumes amsmath package installed
\usepackage{xcolor}
\usepackage{colortbl}
\usepackage{array,colortbl,ctable}
\newcolumntype{P}[1]{>{\centering\arraybackslash}m{#1}}
\usepackage{multirow}

\title{BOOTSTRAPPING GRAPH CONVOLUTIONAL NEURAL NETWORKS FOR AUTISM SPECTRUM DISORDER CLASSIFICATION}

\name{Rushil Anirudh\thanks{This work was performed under the auspices of the U.S. Department of Energy by Lawrence Livermore National Laboratory under Contract DE-AC52-07NA27344.} and Jayaraman J. Thiagarajan}
\address{Lawrence Livermore National Laboratory \\ Email: \{anirudh1, jjayaram\}@llnl.gov}

\begin{document}
%\ninept
%
\maketitle
\begin{abstract}
Using predictive models to identify patterns that can act as biomarkers for different neuropathoglogical conditions is becoming highly prevalent. In this paper, we consider the problem of Autism Spectrum Disorder (ASD) classification where previous work has shown that it can be beneficial to incorporate a wide variety of meta features, such as socio-cultural traits, into predictive modeling. A graph-based approach naturally suits these scenarios, where a contextual graph captures traits that characterize a population, while the specific brain activity patterns are utilized as a multivariate signal at the nodes. Graph neural networks have shown improvements in inferencing with graph-structured data. Though the underlying graph strongly dictates the overall performance, there exists no systematic way of choosing an appropriate graph in practice, thus making predictive models non-robust. To address this, we propose a bootstrapped version of graph convolutional neural networks (G-CNNs) that utilizes an ensemble of weakly trained G-CNNs, and reduce the sensitivity of models on the choice of graph construction. We demonstrate its effectiveness on the challenging Autism Brain Imaging Data Exchange (ABIDE) dataset and show that our approach improves upon recently proposed graph-based neural networks. We also show that our method remains more robust to noisy graphs.
\end{abstract}

\section{Introduction}
\label{sec:intro}
Modeling relationships between functional or structural regions in the brain is a significant step towards understanding, diagnosing and eventually treating a gamut of neurological conditions including epilepsy, stroke, and autism. A variety of sensing mechanisms, such as functional-MRI, Electroencephalography (EEG) and Electrocorticography (ECoG), are commonly adopted to uncover patterns in both brain structure and function. In particular, the resting state fMRI \cite{Kelly2008} has been proven effective in identifying diagnostic biomarkers for mental health conditions such as the Alzheimer disease \cite{Chen2011} and autism \cite{Plitt2015}. At the core of these neuropathology studies are predictive models that map variations in brain functionality, obtained as time-series measurements in regions of interest, to clinical scores. For example, the Autism Brain Imaging Data Exchange (ABIDE) is a collaborative effort \cite{ABIDEpaper2014}, which seeks to build a \emph{data-driven} approach for autism diagnosis. Further, several published studies have reported that predictive models can reveal patterns in brain activity that act as effective biomarkers for classifying patients with mental illness \cite{Plitt2015}. 

%\begin{figure*}[!htb]
%	\centering
%	\includegraphics[clip=true, trim=80 170 80 150,width = 0.95 \linewidth]{figs/arch.pdf}
%	\caption{\small{A generic architecture for machine learning driven neuropathology studies. In this paper, we investigate approaches for incorporating patient similarity into predictive modeling.}}
%	\label{fig:arch}
%	\vspace{-10pt}
%\end{figure*}

%Figure \ref{fig:arch} illustrates a generic pipeline used in these studies. Given the rest state fMRI measurements, the functional connectivities between the different regions of the brain can be estimated. Though the network can be constructed using the individual voxels, it is common practice to extract regions of interest (ROI) based on pre-defined atlases or the correlation structure in the data \cite{Calhoun2001}. In addition to making the analysis more interpretable, this process enables dimensionality reduction by allowing the use of a single representative time-series for each region \cite{Behzadi2007}. Building predictive models requires the use of appropriate features for each subject, whose brain activity is represented as a multivariate time series. While exploiting the statistics of features, e.g. covariance structure of the multivariate time series data, is critical to building effective models, it can be highly beneficial to utilize other non-imaging characteristics shared across subjects from a larger population. 

Graphs provide a natural framework to analyze relationships in a population of patients. The advances in graph signal processing and the generalization of deep neural networks to arbitrarily structured data, make graph models an attractive solution for autism diagnosis. In addition to exploiting correlations in imaging features (e.g. fMRI), the graphs could include a wide-range of non-imaging, contextual features based on more general characteristics of the subjects, including geographical, socio-cultural or gender based features. Recently it has been shown that including such information with a graph convolutional neural network (G-CNN) \cite{Defferrard2016GCNN,Kipf2016GCNN}, can improve classification performance \cite{Parisot2017Spectral}. Despite the applicability of graph-based models in clinical prediction, it is critical to note that the choice of graph construction is crucial to the success of this pipeline -- a low-quality graph can lead sub-optimal models, sometimes even worse than simpler methods that do not utilize the meta information. However, there is currently no definitive strategy on constructing reliable population graphs, and as a result there is a need to build robustness into graph based predictive models, so that they can work with a wider variety of graphs. 
\vspace{10pt}

\noindent \textbf{Proposed Work:} In this paper, we address these issues with a new approach to graph-based predictive modeling, which relies on generating an ensemble of population graphs. First, using a bootstrapping approach to design graph ensembles allows our predictive model to better explore connections between subjects in a large population graph that are not captured by simple heuristics. Second, G-CNNs provide a powerful computing framework to make inferences on graphs, by treating the subject-specific image features as a function, $f: \mathcal{V} \mapsto \mathbb{R}^N$, defined at the nodes of the population graph. We consider a variety of graphs -- (a) feature based graph, $\mathcal{G}_0$, weighted by sex and site \cite{Parisot2017Spectral} (b) a noisy version of $\mathcal{G}_0$, with 30\% of its edges dropped, and (c) a Na\"ive graph, where the adjacency matrix is identity; which is equivalent to using the imaging features alone. Our results show that in every case the proposed bootstrapped G-CNN approach improves classification performance, and provides robustness to the choice of the graph. This simple approach, improves the state of the art classification performance on the ABIDE dataset and also reduces sensitivity of the resulting model to the graph construction step. Consequently, even non-experts can design population graphs, which, with bootstrapping, can perform on par with more sophisticated graph construction strategies.

\section{Predictive Modeling with Ensembles}
\label{sec:approach}
%More specifically, we describe the feature extraction procedure and the different strategies adopted for constructing population graphs. In the next section, we will present the predictive modeling algorithm, based on graph CNNs, that incorporates both the extracted features and information from the population graph.

\begin{figure}[!t]
	\centering
	\includegraphics[clip=True,trim=10 20 10 20,height=0.9\linewidth]{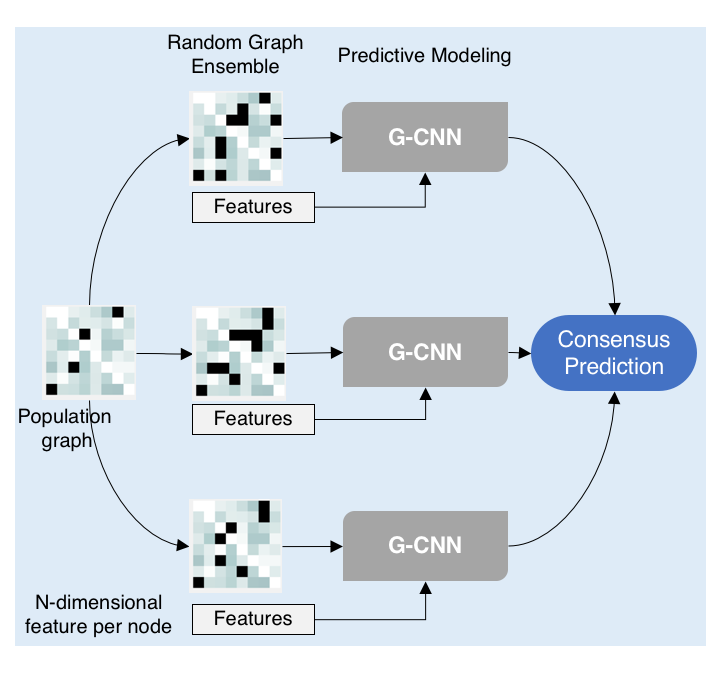}
	\caption{\small{An overview of the proposed approach for predictive modeling with non-imaging features encoded as a population graph and imaging features used as functions at the nodes. We construct a randomized ensemble of population graphs, employ graph CNNs and utilize a consensus strategy to perform the actual classification.}\vspace{-0.22in}}
	\label{fig:app}
\end{figure}

In this section, we describe the proposed approach for predictive modeling to classify subjects with autism. Figure \ref{fig:app} illustrates an overview of the proposed approach for ASD classification. As it can be observed, the pipeline requires an initial population graph and the features at each node (i.e. subject) in the graph as inputs. Subsequently, we create an ensemble of randomized graph realizations and invoke the training of a graph CNN model for every realization. The output layer of these neural networks implement the \textit{softmax} function, which computes the probabilities for class association for each node. Finally, the consensus module fuses the decisions from the ensemble to obtain the final class label. Next we outline the process of graph construction and training strategy for the proposed approach.

\subsection{Population Graph Construction}
A classifier trained on the imaging features alone fails to incorporate contextual non-imaging/meta information that can be critical to discriminate between different classes. For example, it is likely that there is discrepancy in some aspects of data collection at different sites, or the gender of the subject is important in generalizing autism spectrum disorder predictors. It is non-trivial to directly incorporate such information into the subject features, but a graph can be a very intuitive way to introduce these relationships into the learning process. 

\noindent An inherent challenge is that the results obtained are directly dictated by a weighted graph defined for the analysis. Consequently, designing appropriate weighted graphs that capture the geometric structure of data is essential for meaningful analysis. In our context, the population graph construction determines how two subjects are connected, so that context information could be shared between them.  We follow the graph construction strategy used in \cite{Parisot2017Spectral}, which uses a combination of imaging features, gender and site information. First the sex-site graph is obtained as follows: if two subjects have the same gender, they are given a score of $s_{sex} =\lambda_1>1$, and $1$ if they are not. Similarly, the subjects were given a score of $s_{site} = \lambda_2>1$, if they were processed at the same site, and $1$ if not. Next, a linear-kernel graph is computed where the edge weights of the graph are the Euclidean dot product or a linear kernel, between connectivity features from two different ROIs, for a given subject. As expected, this graph does not provide any additional information because it is directly based on the features that were defined at the nodes.

\subsection{Graph Convolutional Neural Networks}
Convolutional neural networks enable extraction of statistical features from structured data, in the form of local stationary patterns, and their aggregation for different semantic analysis tasks, e.g. image recognition or activity analysis. When the signal of interest does not lie on a regular domain, for example graphs, generalizing CNNs is particuarly challenging due to the presence of convolution and pooling operators, typically defined on regular grids. 

\noindent Existing work on generalizing CNNs to graphs can be categorized into \textit{spectral} approaches~\cite{bruna2013spectral, defferrard2016convolutional}, which operate on an explicit spectral representation of the graphs, and \textit{non-spectral} approaches that define convolutions directly on the graphs using spatial neighborhoods~\cite{duvenaud2015convolutional, niepert2016learning}. Spectral approaches, as the name suggests operate using the spectral representation of graph signals, defined using the eigenvectors of the graph Laplacian. For example, in \cite{bruna2013spectral}, convolutions are realized as multiplications in the graph Fourier domain, However, since the filters cannot be spatially localized on arbitrary graphs, this relies on explicit computation of the spectrum based on matrix inversion. Consequently, special families of spatially localized filters have been considered. Examples include the localization technique in \cite{henaff2015deep}, and Chebyshev polynomial expansion based localization in \cite{defferrard2016convolutional}. Building upon this idea, Kipf and Welling \cite{kipf2016semi} introduced graph convolutional neural networks (G-CNN) using localized first-order approximation of spectral graph convolutions, wherein the filters operate within an one-step neighborhood, thus making it scalable to even large networks.

%In spectral approaches, this challenge is alleviated by switching to the spectral domain, where the convolution operations can be viewed as simple multiplications. In general, there exists no mathematical definition for the translation operation on graphs. However, a spectral domain approach defines the localization operator on graphs via convolution with a Kronecker delta signal. However, localizing a filter in the spectral domain requires the computation of the graph Fourier transform and hence translations on graphs are computationally expensive. We refer the reader to \cite{Kipf2016GCNN, Defferrard2016GCNN} for a detailed description of graph convolutional neural network architectures. 

\subsection{Ensemble Learning}
As described in the previous section, population graphs provide a convenient way to incorporate non-imaging features into the predictive modeling framework, where the connectivity features from fMRI data are used as a function on the graph. While G-CNN can automatically infer spectral filters for achieving  discrimination across different classes, the sensitivity of its performance to the choice of the population graph is not straightforward to understand. Consequently, debugging and fine-tuning these networks can be quite challenging. A classical approach to building robust predictive models is to infer an ensemble of weak learners from data and then fuse the decisions using a consensus strategy. The need for ensemble models in supervised learning has been well established \cite{Dietterich2000}. A variety of bootstrapping techniques have been developed, wherein multiple ``weak'' models, that focus on subsets of the data or different aspects of the task, are inferred and finally fused for effective prediction. The intuition behind the success of this approach is that different models may have a similar training error when learned using a subset of training samples, but their performance on test data can be different since they optimized for different regions of the input space. 

\vspace{10pt}

\noindent We employ a similar intuition to building models with population graphs, wherein the quality of different similarity metrics for graph construction can lead to vastly different predictive models. More specifically, starting with a population graph, we create an ensemble of graphs, $\{\mathcal{G}_p\}_{p=1}^P$, by dropping out a pre-defined factor of edges randomly. In this paper, we use a uniform random distribution for the dropout, though more sophisticated weighted distributions could be used. For each of the graphs in the ensemble, we build a G-CNN model with the connectivity features as the $N-$dimensional multivariate function at each node. The output of each of the networks is a softmax function for each node, indicating the probability for the subject to be affected by ASD. Note that, unlike conventional ensemble learners, we do not subsample data, but only drop edges from the population graph. Conceptually, this is similar to the idea of using multiple attention heads in the recently successful attention models in deep learning \cite{shanthamallu2018attention}. Given the logits from all the weak learners, we employ simple consensus strategies such as averaging of the probabilities estimated by each of the G-CNNs for a test subject. As we will show in our experiments, the proposed ensemble approach boosts the performance of all the population graph construction strategies considered. 

\begin{figure}[!htb]
\centering
\includegraphics[clip=True,trim=0 0 0 30,height=0.7\linewidth]{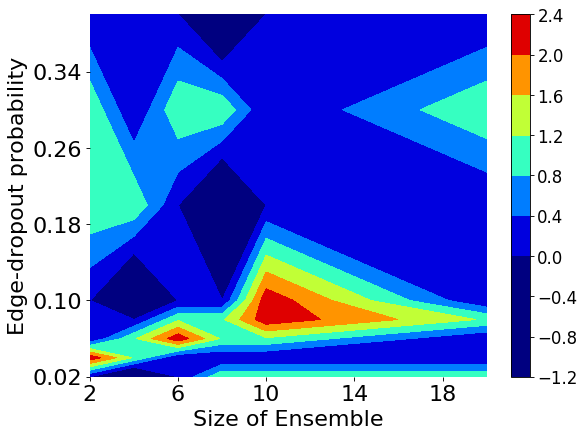}
\caption{\small{\textbf{Performance improvement with Bootstrapping:} An exhaustive hyper-parameter search demonstrating the improvement in performance obtained by size of the graph ensemble, and edge dropout probability. We show results for a particular test/train split here.}}
\label{fig:hyper}
\vspace{-30pt}
\end{figure}

\noindent The two main hyper-parameters in our approach are the size of the ensemble, and the edge-dropout probability. In general we observe that a large ensemble (around 20 graphs) works effectively, while a dropout probability of around 0.2-0.3 improves performance consistently. Figure \ref{fig:hyper} shows improvement in performance for a particular train/test split in the ABIDE dataset used in our experiments. As it is seen, there are several combinations that lead to an improvement in performance, while not requiring any additional information.

%\section{Predictive Modeling: Randomized Ensemble of G-CNNs}
%\label{sec:graph}
%\input{graph.tex}

\section{Experiments}
\label{sec:results}
In this section, we describe our experiments to evaluate the performance of the proposed ensemble G-CNN approach in ASD classification. 

\noindent \textbf{The ABIDE dataset:} We present our results on the Autism Brain Imaging Data Exchange (ABIDE) dataset \cite{ABIDEpaper2014} that contains resting state fMRI data (rs-fMRI) for 1112 patients, as part of the preprocessed connectomes project \footnote{\url{http://preprocessed-connectomes-project.org/abide/}}. The pre-processing includes slice-timing correction, motion correction, and intensity normalization, depending on the pipeline used. We follow the same preprocessing pipeline (C-PAC) and atlases (Harvard-Oxford) as described in \cite{Abraham2017NeuroImage}, in order to facilitate easy comparison -- this resulted in a dataset with 872 of the initial 1112 patients available, from 20 different sites. The task is to diagnose a patient as being of two classes -- Autism Spectrum Disorder (ASD) or Typical Control (TC). The dataset along with different pre-processing strategies are available via the nilearn python package \footnote{\url{https://nilearn.github.io/introduction.html}}.
%These labels are available separately \footnote{Column named DX\textunderscore GROUP in \url{https://s3.amazonaws.com/fcp-indi/data/Projects/ABIDE_Initiative/Phenotypic_V1_0b_preprocessed1.csv}}.

The resulting data per subject consists of the mean time series obtained from the rs-fMRI for each Region of Interest (ROI). In total, there are 111 ROIs for the HO atlas considered here, resulting in the data, for the $i^{th}$ subject, of size $\mathbb{R}^{111 \times T}$ matrix, where $T$ is the total number of slices in the fMRI measurement. We use the same 10-fold cross validation as in \cite{Parisot2017Spectral} of the ABIDE dataset, using their publicly released code \footnote{\url{https://github.com/parisots/population-gcn}}.  
\vspace{5pt}

\noindent \textbf{Training details and parameters:} For a fair comparison, we use the same G-CNN set up used in \cite{Parisot2017Spectral}, which utilizes the original implementation from \cite{Kipf2016GCNN}. We constructed a fully graph-convolutional network, with 3 layers of 16 units each; a learning rate of 0.005 and dropout of 0.3. We used Chebychev polynomials upto degree 3 to approximate the Graph Fourier Transform in the G-CNN in all our experiments. We also use recursive feature elimination to reduce the feature space to $2000$ most important features. All our models, including the ensembles are trained for $200$ epochs each.
\vspace{10pt}

\noindent \textbf{Ensemble G-CNN:} For a given graph $\mathcal{G}$, we generate $P$ new graphs such that the new set of graphs $\{\mathcal{G}_0, \mathcal{G}_1 \dots \mathcal{G}_P\}$ were obtained by randomly dropping $30-40\%$ of the edges of $\mathcal{G}$. The predictions from each model obtained using the ensemble are fused as follows: we take the mean of the predictions and assign the class as the one with the largest value. In general, each member of the random ensemble behaves as a \emph{weak learner}, where the performance for each individual network is suboptimal, but the consensus decision is better than the state-of-the-art. In general we observe that for \emph{reliable} graphs, edge-drop probability of $0.25-0.3$ was effective, while for noisier or sparser graphs even small perturbations of $0.05$ were found to be effective in boosting performance. In both cases we observed better performance with a large ensemble of around $20-25$ random graphs. A hyper-parameter search comparing the number of graphs and edge drop probability for a particular test/train split is shown in figure \ref{fig:hyper}.

\begin{table}[t]
	\centering
	\renewcommand*{\arraystretch}{1.3}
	\begin{tabular}{|c|c|c|}
		\hline
		\cellcolor{gray!15}\textbf{Population Graph}                     & \cellcolor{gray!15}\textbf{Predictive Model}                      & \cellcolor{gray!15}\textbf{Accuracy} \\ \hline \hline
		-                                           & Linear SVM\cite{Abraham2017NeuroImage} & 66.8              \\ \hline \hline
		\multirow{2}{*}{$\mathcal{G}$}                & G-CNN  \cite{Parisot2017Spectral}              & 69.50             \\ \cline{2-3} 
		& Proposed                                       & \textbf{70.86}    \\ \hline \hline
		\multirow{2}{*}{Na\"ive Graph} & G-CNN \cite{Parisot2017Spectral}           & 66.93             \\ \cline{2-3} 
		& Proposed                                       & \textbf{67.85}    \\ \hline \hline
		\multirow{2}{*}{Noisy $\mathcal{G}$}          & G-CNN    \cite{Parisot2017Spectral}      & 66.35             \\ \cline{2-3} 
		& Proposed                                       & \textbf{67.39}    \\ \hline
	\end{tabular}
  \caption{Classification Accuracy for the ABIDE dataset (\cite{ABIDEpaper2014}) using the proposed approach. For comparison, we report the state-of-the-art results obtained using linear SVM  and G-CNN . $\mathcal{G}$ corresponds to the graph proposed in \cite{Parisot2017Spectral}, that uses the patient sex and hospital location information, along with the f-MRI features. Na\"ive Graph refers to using an identity matrix as the adjacency, assuming graph information is not available. Finally noisy $\mathcal{G}$ refers to the case when we drop 30\% of the edges from the graph.}
  \label{tab:results}
\end{table}
\subsection{Classification Results}
The classification accuracy obtained by using the bootstrapping approach is shown in table \ref{tab:results}. A few observations are important to note -- There is a clear and obvious advantage in using the graph based approach to classifying populations. We out perform recent state of the art method \cite{Parisot2017Spectral} for ASD classification by nearly $1.5$ percentage points, without any additional information, and using the exact same training protocol with graph CNNs. This advantage is also seen in cases where the graphs are noisy, and also when no graph information is available at all. In both cases, we see the proposed approach demonstrates robustness to the changes in the graph. Secondly, the difference in performance for different graphs illustrates the importance of graph construction, further emphasizing the need for robust training strategies such as those proposed in this paper. Finally, it can be observed that the best performing split has a consistently high performance across different kinds of graphs and only marginally better than the baseline, perhaps indicating that the connectivity features dictate the performance in that case.
\section{Discussion and Future Work}
\label{sec:disc}
In this paper we presented a training strategy for graph convolutional neural networks (G-CNNs), that have been recently proposed as a promising solution to the node classification problem in graph structured data. We focused particularly on autism spectrum disorder classification using time series extracted from resting state fMRI. While recent graph-based predictive models have shown that incorporating contextual non-imaging features such as patient sex, location of the scan etc. can improve performance, they end up being extremely sensitive to the choice of the population graph. To circumvent this challenge, we propose to use bootstrapping as a way to reduce the sensitivity of the initial graph construction step, by generating multiple random graphs from the initial population graph. We train a G-CNN for each randomized graph, and fuse their predictions at the end. These individual predictive models behave as weak learners, that can be aggregated to produce superior classification performance. The proposed work opens several new avenues of future work including (a) pursuing a theoretical justification behind the improved performance from randomized ensembles, (b) extending these ideas by using an ensemble of random binary graphs which are very cheap to construct, and (c) training the ensemble of networks together.

\small
\bibliographystyle{IEEEbib}
\bibliography{refs,refs-graphs}
\section*{Disclaimer}
This document was prepared as an account of work sponsored by an agency of the United States government.
Neither the United States government nor Lawrence Livermore National Security, LLC, nor any of their
employees makes any warranty, expressed or implied, or assumes any legal liability or responsibility for the
accuracy, completeness, or usefulness of any information, apparatus, product, or process disclosed, or represents
that its use would not infringe privately owned rights. Reference herein to any specific commercial product,
process, or service by trade name, trademark, manufacturer, or otherwise does not necessarily constitute or
imply its endorsement, recommendation, or favoring by the United States government or Lawrence Livermore
National Security, LLC. The views and opinions of authors expressed herein do not necessarily state or reflect
those of the United States government or Lawrence Livermore National Security, LLC, and shall not be used for
advertising or product endorsement purposes.

\end{document}